\documentclass[11pt]{article}

\usepackage[margin=1in]{geometry}

\usepackage{fontspec}
\usepackage{xeCJK}

\setCJKsansfont[
    BoldFont = FandolHei-Bold.otf
]{FandolHei-Regular.otf}

\setCJKmonofont{FandolFang-Regular.otf}

\newCJKfontfamily\cjkcodefont[
    Scale=0.80,
    AutoFakeBold=false
]{FandolSong-Regular.otf}

\newcommand{\cjktok}[1]{{\cjkcodefont\mdseries\upshape #1}}

\usepackage{algorithm}
\usepackage{algpseudocode}
\usepackage{amsmath, amssymb}
\usepackage{array}
\usepackage{booktabs}
\usepackage{caption}
\usepackage{csquotes}
\usepackage{graphicx}
\usepackage[hidelinks]{hyperref}
\hypersetup{
    pdftitle={Dynamic-Programming-Guided Hierarchical BPE and Empirical Analysis of Vocabulary Pruning},
    pdfauthor={Kenny Shao},
    pdfsubject={Byte Pair Encoding and tokenizer vocabulary optimization},
    pdfkeywords={Byte Pair Encoding, BPE, tokenization, vocabulary pruning,
dynamic programming, hierarchical tokenization}
}
\usepackage{multirow}
\usepackage{makecell}
\usepackage{placeins}
\usepackage{pgfplots}
\usepackage{tabularx}
\usepgfplotslibrary{groupplots}
\pgfplotsset{compat=1.18}

\pgfplotsset{
    standardbpe/.style={
        draw=blue!80!black,
        fill=blue!30,
        every node near coord/.append style={
            text=blue!80!black
        }
    },
    prunedbpe/.style={
        draw=purple!80!black,
        fill=purple!25,
        every node near coord/.append style={
            text=purple!80!black
        }
    }
}

\title{Dynamic-Programming-Guided Hierarchical BPE and Empirical Analysis of Vocabulary Pruning}

\author{
Kenny Shao \\
\texttt{kennyshao0919@gmail.com}
}

\date{}

\begin{document}

\maketitle

\begin{abstract}
Byte Pair Encoding (BPE) constructs vocabularies through greedy pair merging, but the resulting merge order does not necessarily allocate a fixed model-visible vocabulary optimally for compression. We propose 
\emph{Dynamic-Programming-Guided Hierarchical BPE} (DH-BPE), a vocabulary-construction
method that combines token exposure under exact minimum-token segmentation with the hierarchical dependencies induced by BPE training. Starting from a modestly overshot BPE candidate vocabulary, DH-BPE uses dynamic programming to measure candidate utility and applies exposure-guided, dependency-aware pruning to select a fixed-size model-visible
vocabulary. We compare DH-BPE against Standard BPE and recent vocabulary-optimization baselines, including Pruned BPE, MinGram, and MinGram-PP, in primary evaluations at 12K and 16K target vocabulary sizes, with an additional 18K evaluation against MinGram only. Across the primary 12K and 16K comparisons, DH-BPE consistently improves aggregate compression
over Standard BPE, Pruned BPE, and MinGram under a shared exact minimum-token
DP encoder. MinGram-PP achieves stronger aggregate compression in the primary
comparisons, but DH-BPE outperforms it at overshoot factors \(f=2.0\) and \(f=3.0\) in 
cross-corpus evaluation; at 12K, MinGram-PP reverses this ordering only with the substantially larger candidate pools at \(f=4.0\) and \(f=5.0\). Qualitative analysis further shows that DH-BPE balances later, more complete BPE merges with reusable subword components, providing a practical approach to improving vocabulary allocation under a
fixed model-visible vocabulary budget.
\end{abstract}

\section{Introduction}
Byte Pair Encoding (BPE) has become one of the most widely used subword tokenization approaches for large language models (LLMs)~\cite{sennrich2016neural}. Its greedy merge procedure is simple, efficient, and capable of constructing compact vocabularies from raw text, but the vocabulary produced during training is not necessarily optimal for final model-visible use. In our previous work on Pruned BPE~\cite{shao2026prunedbpe}, we showed that some tokens created during BPE training are useful primarily as intermediate components of later merges while appearing only rarely in the final encoded corpus. Pruned BPE addresses this mismatch through post-training visibility pruning and token reallocation, allowing low-exposure tokens to remain internally available when required by later merges while reallocating visible vocabulary slots to more useful tokens. This naturally raises a further question: can the usefulness of BPE tokens be evaluated more directly when deciding which learned tokens should remain model-visible?

A useful perspective comes from minimum-token segmentation~\cite{schmidt2024tokenization,raj2025everytoken}. Given a fixed vocabulary, dynamic programming can determine an exact
segmentation of a sequence that minimizes the number of tokens required to represent it.
Such a segmentation provides more than an encoding result: it also reveals which vocabulary entries are actually selected when candidate tokens compete under the objective of minimizing encoded length. Tokens that repeatedly appear in these minimum-token segmentations can therefore provide a direct signal of their utility to compression. At the same time, the sequential merge process of standard BPE naturally creates a hierarchical dependency structure among learned tokens. Each merged token is formed from two previously available components and can therefore be recursively traced to lower-level tokens. This structure provides additional information that can be exploited during vocabulary pruning, allowing pruning decisions to account for the dependencies among learned tokens rather than treating them independently.

Motivated by these observations, we propose
\emph{Dynamic-Programming-Guided Hierarchical BPE} (DH-BPE). DH-BPE uses token exposure under exact minimum-token segmentation together with the hierarchical structure created by BPE merges to select the final model-visible vocabulary. Starting from a BPE vocabulary trained beyond the desired final size according to a specified overshoot factor, the method applies minimum-token dynamic programming to the training corpus, measures how frequently candidate tokens are selected in the resulting segmentations, and uses these exposure statistics to guide hierarchical vocabulary pruning. During the pruning process, lower-level tokens required to preserve the BPE dependency structure of retained higher-level candidates are protected from premature removal, even if they are not ultimately selected for the final model-visible vocabulary. After pruning is complete, only the selected model-visible vocabulary is retained for encoding. DH-BPE therefore leaves the standard BPE merge-learning procedure unchanged and instead introduces a post-training
mechanism for reducing the learned vocabulary to a fixed target size.

The contributions of this work are threefold. First, we introduce DH-BPE, a vocabulary-pruning framework that combines exposure measured under exact minimum-token segmentation with dependency-aware pruning of a BPE-derived vocabulary. Second, we systematically investigate important design choices of the proposed method, including two-space token protection, the BPE vocabulary overshoot factor, and the pruning ratio, and evaluate how these choices affect tokenizer compression. Third, we conduct controlled same-corpus and cross-corpus evaluations together with qualitative analysis of retained vocabulary entries, providing insight into how DH-BPE distributes limited vocabulary capacity between longer corpus-specific tokens and shorter reusable components and how these choices affect generalization across corpora.

\section{Related Work}
Standard BPE constructs a subword vocabulary by repeatedly merging the most frequent adjacent token pair until a target vocabulary size is reached~\cite{sennrich2016neural}. Because every newly created token is formed from previously available tokens, this process also induces dependencies among learned tokens. Prior work relevant to DH-BPE addresses two closely related questions: how to segment text under a fixed vocabulary, and how to construct
or refine the vocabulary itself. The latter includes compression-oriented vocabulary construction, BPE vocabulary pruning, post-training adaptation, and structure-aware vocabulary modification.

Prior work has explored alternative segmentation procedures for a fixed subword vocabulary. He et al.~\cite{he2020dpe} proposed Dynamic Programming Encoding (DPE), which uses dynamic programming to find a maximum-posterior segmentation under a probabilistic model. Raj S et al.~\cite{raj2025everytoken} instead study compression-oriented inference for a fixed BPE vocabulary, using dynamic programming to find a segmentation with the minimum number of tokens. Related studies have more broadly examined the separation between vocabulary construction and tokenizer inference. Uzan et al.~\cite{uzan2024greed} compare alternative inference procedures for fixed subword vocabularies, while Sawada and Goyal~\cite{sawada2025train} study BPE inference without relying on the original merge list, including greedy and exact compression-oriented procedures. These works focus on segmentation or inference for an already available vocabulary rather than using token exposure under minimum-token segmentation to guide BPE vocabulary pruning.

Minimum-token objectives and other compression-oriented approaches have also
been used during vocabulary construction. Schmidt et al.~\cite{schmidt2024tokenization} introduced PathPiece, which finds an exact minimum-token segmentation for a given
vocabulary and constructs vocabularies in a top-down manner by iteratively
removing candidates whose omission is estimated to cause the smallest increase
in corpus token count. Other work approaches compression-oriented vocabulary
construction through more global optimization. GreedTok~\cite{lim2025partition}
formulates fixed-size vocabulary selection as a combinatorial optimization
problem, ConvexTok~\cite{tempus2026convextok} uses a linear-programming relaxation, and ToaST~\cite{schmidt2026toast} combines split-tree inference with integer- and linear-programming-based vocabulary selection. These methods provide alternative ways to optimize token count during vocabulary construction, but they do not combine minimum-token exposure with the dependency hierarchy induced by BPE training.

A separate line of work prunes or refines BPE-derived vocabularies.
SaGe~\cite{yehezkel2023sage} begins with an oversized candidate vocabulary
obtained by over-applying a base tokenizer such as BPE and iteratively removes
tokens according to a contextual Skip-Gram ablation signal. BPE vocabulary
trimming~\cite{cognetta2024analysis} instead removes rare subwords after BPE
training and represents them through their component subwords.
PickyBPE~\cite{chizhov2024picky} modifies vocabulary membership during BPE
training using an Intersection-over-Self (IoS) statistic based on token and
pair frequencies to identify intermediate tokens that become largely redundant
after participating in larger merges. More recently, LiteToken~\cite{sun2026litetoken} directly targets intermediate merge residues---tokens that are frequent enough to be created during BPE training but are later mostly absorbed into larger tokens and rarely emitted.
These approaches share the general goal of reducing inefficient or underused BPE vocabulary entries, but their selection signals differ from DH-BPE's exposure measured under exact
minimum-token segmentation.

Post-training tokenizer adaptation provides a related form of vocabulary reallocation. AdaptBPE~\cite{liyanage2026adaptbpe} revises an existing BPE merge list under
a fixed merge budget, replacing low-utility merges with candidates favored by
an adaptation corpus. Didenko~\cite{didenko2026writing} similarly studies fixed-size byte-level BPE adaptation for target writing systems while explicitly preserving
merge-graph reachability during token insertion. These methods primarily address adaptation of an existing tokenizer to a new domain, language, or writing system, whereas DH-BPE uses the tokenizer-training corpus itself to construct a compression-oriented vocabulary.

Other work explicitly exploits the dependency structure induced by BPE merges.
BPE-knockout~\cite{bauwens2024knockout} represents a pre-existing BPE
tokenizer as a merge graph and removes undesirable tokens while rewriting
dependent merges so that higher-level tokens are not unintentionally lost.
Its pruning signal is based primarily on morphological semi-supervision rather
than token-count compression. ReBPE~\cite{bauwens2026rebpe} extends this
approach by alternating pruning with a reification procedure that mines the
internal structure of the modified tokenizer to introduce new merges.
Purason et al.~\cite{purason2026teaching} instead restrict vocabulary pruning
to tokens that are currently leaves in the BPE merge graph, thereby preserving
downstream merge dependencies. These approaches demonstrate that BPE
dependency structure can be useful during vocabulary modification, but they do
not combine that structure with token exposure measured under exact
minimum-token segmentation.

Table~\ref{tab:related-work-comparison} summarizes a selected set of methods
most directly related to DH-BPE in terms of minimum-token-based vocabulary
construction, BPE vocabulary pruning or reallocation, and explicit use of BPE
structure. Broader optimization, inference-only, and more specialized
adaptation methods are discussed in the text above. Pruned BPE and the two
MinGram variants are omitted from the table because they are discussed
separately below.

\begin{table}[!ht]
\centering
\scriptsize
\setlength{\tabcolsep}{2.5pt}
\renewcommand{\arraystretch}{1.12}

\begin{tabularx}{\textwidth}{@{}
    p{1.45cm}
    >{\raggedright\arraybackslash}X
    >{\raggedright\arraybackslash}X
    p{1.65cm}
    >{\raggedright\arraybackslash}X
    >{\raggedright\arraybackslash}X
@{}}

\toprule

\textbf{Method} &
\textbf{Vocabulary modification} &
\textbf{Segmentation / optimization} &
\textbf{BPE}\newline \textbf{dependency} &
\textbf{Selection or removal signal} &
\textbf{Primary objective} \\

\midrule

PathPiece &
Top-down pruning from a large seed &
Exact minimum-token DP &
No explicit use &
Estimated token-count increase after omission &
Token-count compression \\

\midrule

SaGe &
Top-down pruning from an oversized seed &
Iterative re-tokenization with contextual scoring &
No explicit use &
Contextual Skip-Gram ablation loss &
Context-aware vocabulary construction \\

\midrule

BPE\newline trimming &
Post-training vocabulary reduction &
BPE encoding with recursive decomposition &
Uses BPE\newline decomposition &
Final token-frequency threshold &
Vocabulary reduction \\

\midrule

PickyBPE &
During BPE training &
Modified BPE training &
No explicit\newline merge graph &
Intersection-over-Self criterion &
Remove redundant intermediate tokens \\

\midrule

LiteToken &
Post-training residue removal &
BPE-based split and re-merge &
Yes &
Intermediate/final usage and neighbor-diversity signals &
Remove underused merge residues \\

\midrule

AdaptBPE &
Fixed-budget merge replacement &
BPE merge adaptation &
Yes &
Adaptation-corpus frequency and compression &
Domain/language adaptation \\

\midrule

BPE-knockout &
Post-training merge removal &
Modified BPE merge graph &
Yes &
Morphological merge utility &
Morphological alignment and pruning \\

\midrule

\textbf{DH-BPE} &
Post-BPE pruning from an overshoot vocabulary &
Exact minimum-token DP &
Yes &
\textbf{Minimum-token exposure with BPE hierarchy} &
Fixed-size vocabulary allocation and compression \\

\bottomrule
\end{tabularx}

\caption{Comparison of DH-BPE with closely related BPE vocabulary-construction
and refinement methods}
\label{tab:related-work-comparison}
\end{table}

\subsection{Pruned BPE}
Our previous work, Pruned BPE~\cite{shao2026prunedbpe}, separates BPE merge construction from model-visible vocabulary allocation. After standard BPE training, it measures \emph{final exposure}, defined by how often a learned token remains in the final Standard BPE encoding of the training corpus. Tokens with insufficient exposure may be designated internal-only, while their model-visible vocabulary positions are reallocated to additional candidates learned through resumed BPE training. The internal-only tokens remain part of the tokenizer's merge structure and are recursively expanded into visible descendants during encoding.

DH-BPE retains the general motivation of allocating a fixed vocabulary more effectively but changes both the evaluation criterion and the role of the BPE hierarchy. Candidate exposure is measured under exact minimum-token segmentation rather than under the final segmentation produced by Standard BPE merge application. This exposure signal is then combined with the hierarchical relationships produced during BPE training to guide vocabulary pruning. The hierarchy is used during vocabulary construction and pruning; after this process is complete, only the selected model-visible vocabulary is required by the DH-BPE encoder. Thus, unlike Pruned BPE, DH-BPE does not require internal-only merge tokens or recursive expansion during final encoding.

\subsection{MinGram and MinGram-PP}
\label{subsec:related-mingram}
MinGram~\cite{land2026mingram} is a minimalist Unigram tokenizer designed to emphasize compression while simplifying conventional Unigram training. It begins with a BPE-derived candidate vocabulary trained beyond the target size, then performs Hard EM using minimum-token paths. Token count is the primary segmentation objective, with learned Unigram scores used to break ties. During each Hard-EM iteration, counts are accumulated only for tokens on the selected minimum-token paths and used to re-estimate the scores. The final vocabulary is
obtained through flat score-based pruning.

MinGram-PP, introduced in the same work~\cite{land2026mingram}, retains the BPE-derived initialization and minimum-path Hard-EM training of MinGram but replaces flat score pruning with PathPiece-style iterative pruning. Tokens are ranked by the increase in corpus token count caused by their deletion, and low-impact candidates are removed until the target vocabulary size is reached. MinGram-PP therefore places greater emphasis on compression during final vocabulary selection, at the cost of substantially more expensive pruning.

MinGram and MinGram-PP are particularly relevant baselines for DH-BPE because all three approaches begin with BPE-derived candidate vocabularies and use minimum-token segmentation during vocabulary construction. However, the resulting minimum-token information is used differently during vocabulary selection. MinGram uses minimum-path token counts to re-estimate a flat set of Unigram scores and ultimately prunes by those scores, whereas 
MinGram-PP selects tokens using iterative token-deletion loss. DH-BPE instead uses token exposure under exact minimum-token segmentation directly as a pruning signal and combines that signal with the hierarchical relationships inherited from BPE training. The resulting approaches therefore provide a useful controlled comparison of different ways to exploit minimum-token information when allocating a fixed vocabulary.

\section{Dynamic-Programming-Guided Hierarchical BPE}
\label{sec:dh-bpe}

\subsection{Minimum-Token Dynamic Programming}
\label{subsec:min-token-dp}
DH-BPE uses exact minimum-token dynamic programming for two related purposes: to measure token exposure during vocabulary construction and to perform final encoding. Given a fixed vocabulary \(V\) and a byte sequence \(x\) of length \(n\), the objective is to find a valid segmentation of \(x\) that contains the minimum possible number of tokens from \(V\). This minimum-token objective has been explored in prior tokenization work~\cite{schmidt2024tokenization}; DH-BPE uses the resulting segmentation as a signal for vocabulary pruning.

Let \(D(i)\) denote the minimum number of tokens required to encode the suffix of \(x\) beginning at byte position \(i\), for \(0\le i\le n\). The base case is \(D(n)=0\), since position \(n\) corresponds to the empty suffix after the final byte and therefore requires no tokens. For each position \(0\le i<n\), let \(A(i)\) denote the set of vocabulary
tokens whose byte representation matches the input beginning at \(i\).
Then \(D(i)\) satisfies the dynamic-programming recurrence

\[
D(i)
=
1+
\min_{t\in A(i)}
D(i+|t|),
\]

where \(|t|\) is the byte length of token \(t\). Since the vocabulary contains all 256 single-byte tokens, a complete segmentation always exists. The DP table is filled from right to left, and the selected tokens are then recovered from left to right to obtain an exact minimum-token segmentation.

Multiple segmentations may contain the same minimum number of tokens. To make both vocabulary analysis and encoding deterministic, DH-BPE breaks such ties by preferring the longer token at the earliest position where the segmentations differ. If competing tokens at that position have the same byte length, the token with the smaller token ID is preferred.

For vocabulary construction, let \(V_{\text{cand}}\) denote the complete overshoot candidate vocabulary used by DH-BPE, formed from the vocabulary produced by standard BPE overshoot training together with the reserved tokens. The DP segmentation is applied to the training corpus using \(V_{\text{cand}}\). As described in Subsection~\ref{subsec:training-evaluation-data}, pretokenization is applied first, and each resulting chunk is segmented independently. Let \(C_{\mathrm{chunk}}\) denote the collection of these chunks. We define the \emph{DP exposure} of a candidate token \(t\) as

\[
E_{\mathrm{DP}}(t)
=
\sum_{x\in C_{\mathrm{chunk}}}
\operatorname{count}
\left(
t,\,
S_{\mathrm{DP}}(x;V_{\text{cand}})
\right),
\]

where \(S_{\mathrm{DP}}(x;V_{\text{cand}})\) denotes the exact minimum-token
segmentation of a chunk \(x\), treated as a byte sequence as defined above,
under \(V_{\text{cand}}\). Thus, \(E_{\mathrm{DP}}(t)\) counts how often \(t\) is actually selected when all candidate tokens compete under the minimum-token objective. Candidates that are never selected receive zero exposure.

The same DP procedure is used after vocabulary construction for encoding. At that stage, however, the input vocabulary is only the final model-visible vocabulary 
\(V_{\text{visible}}\). The BPE dependency information used during vocabulary construction is not required by the encoder, and no internal-only vocabulary is used during encoding.

\subsection{Hierarchical Vocabulary Pruning}
\label{subsec:hierarchical-pruning}
Standard BPE creates tokens sequentially through binary merges. If a learned token \(t_i\) is produced by merging two tokens \(t_{\mathrm{left}}\) and \(t_{\mathrm{right}}\), both components necessarily exist before \(t_i\) is created. Consequently, the learned BPE vocabulary contains an implicit hierarchy whose creation order respects its dependency relationships. DH-BPE uses this structure together with the DP exposure defined in Subsection~\ref{subsec:min-token-dp} rather than treating the overshoot vocabulary as an unordered collection of independent candidates.

Let \(B=(t_1,t_2,\ldots,t_m)\) denote the learned candidates in their original BPE merge order. Given the exposure threshold \(\tau_r\) corresponding to pruning ratio \(r\), 
a candidate satisfies the DP-exposure criterion when

\begin{equation}
E_{\mathrm{DP}}(t_i)\ge\tau_r.
\label{eq:eligibility-criterion}
\end{equation}

Importantly, DH-BPE uses the threshold as an eligibility criterion rather than ranking eligible candidates by the magnitude of their DP exposure. Eligible candidates are considered in their original BPE merge order until the required number of learned visible tokens has been selected. In this way, DP exposure determines whether a candidate has demonstrated sufficient utility under minimum-token segmentation, while the BPE construction order provides the hierarchical organization for vocabulary selection.

Low-exposure tokens encountered within the retained BPE prefix are not selected for final model visibility. During vocabulary construction, however, they are temporarily retained so that the merge dependencies of later selected tokens remain intact. More precisely, after the final set of visible learned tokens has been selected, let \(t_j\) be the latest-created selected token. The construction-time internal set is

\[
V_{\text{internal}}
=
\{\,t_i\in B :
i\le j,\;
t_i\notin V_{\text{learn}}\,\},
\]

where \(V_{\text{learn}}\) is the set of selected learned visible tokens. Retaining the prefix through \(t_j\) conservatively preserves all earlier BPE components on which selected higher-level tokens may depend.

This construction-time internal set should not be confused with the internal-only vocabulary of Pruned BPE~\cite{shao2026prunedbpe}. In Pruned BPE, internal-only tokens remain part of the tokenizer's merge structure and are expanded during encoding. In DH-BPE, \(V_{\text{internal}}\) is used only to preserve and remap the BPE hierarchy during vocabulary construction. After pruning is complete, only \(V_{\text{visible}}\) is supplied to the minimum-token encoder.

\subsection{Protection of Two-Space Tokens}
\label{subsec:two-space-protection}

A direct exposure threshold can undesirably remove learned tokens associated with repeated whitespace. Under the pretokenization procedure described in Subsection~\ref{subsec:training-evaluation-data}, exactly one ordinary ASCII space may be attached to a following word-like or code-like chunk, whereas runs of two or more consecutive whitespace characters are emitted as independent whitespace chunks. Because BPE training is applied independently within each pretokenized chunk, any learned token beginning with two consecutive ASCII spaces is necessarily a whitespace-only token and cannot extend into adjacent textual content.

To make the pruning procedure more robust to the removal of such repeated-whitespace tokens, DH-BPE protects learned tokens whose byte representation begins with two consecutive ASCII spaces. Let
\[
P =
\{\,t\in B :
t \text{ begins with two consecutive ASCII spaces}\,\}.
\]

Tokens in \(P\) are exempt from the DP-exposure criterion in
Equation~\eqref{eq:eligibility-criterion} and are retained in the
model-visible vocabulary regardless of whether
\(E_{\mathrm{DP}}(t)\ge\tau_r\). All remaining learned candidates are
evaluated normally according to Equation~\eqref{eq:eligibility-criterion}
and, if eligible, retain their original BPE merge order as described in
Subsection~\ref{subsec:hierarchical-pruning}.

If the final vocabulary requires \(N_{\mathrm{required}}\) learned visible
tokens, as defined in Equation~\eqref{eq:n-required} in the next subsection, the 
protected set consumes \(|P|\) of these positions. DH-BPE therefore selects
\(N_{\mathrm{required}}-|P|\) additional eligible non-protected candidates
in original BPE merge order and combines them with \(P\). Protection affects
only learned tokens; the 256 byte-level tokens and reserved tokens are fixed
independently of the pruning process.

The effect of this protection rule is evaluated separately in 
Subsection~\ref{subsec:two-space-experiment}. This ablation allows us to distinguish the effect of hierarchy-aware DP pruning itself from the additional treatment of 
repeated-whitespace tokens.

\subsection{Overshoot Factor and Pruning Ratio}
\label{subsec:overshoot-pruning-ratio}

DH-BPE begins with a BPE vocabulary larger than the desired final model-visible vocabulary. Let \(N_{\text{visible}}\) denote the target visible vocabulary size, 
\(N_{\text{reserved}}\) the number of reserved tokens, and \(N_{\text{byte}}=256\) the number of byte-level tokens. The number of learned tokens required in the final vocabulary is

\begin{equation}
N_{\text{required}}
=
N_{\text{visible}}
-
N_{\text{reserved}}
-
N_{\text{byte}}.
\label{eq:n-required}
\end{equation}

Given an overshoot factor \(f>1\), standard BPE is trained until the ordinary vocabulary reaches

\begin{equation}
N_{\text{train}}
=
N_{\text{byte}}
+
\left\lceil fN_{\text{required}}\right\rceil.
\label{eq:n-train}
\end{equation}

The \(N_{\text{reserved}}\) reserved tokens are then added separately to form the complete overshoot candidate vocabulary \(V_{\text{cand}}\). 

The overshoot factor therefore controls how many learned BPE candidates are included in the overshoot vocabulary \(V_{\text{cand}}\). A larger \(f\) exposes the pruning procedure to more later-created BPE tokens, but also increases the number of candidates that must be evaluated. If the candidate pool does not contain enough eligible tokens to fill the final visible vocabulary, BPE training must continue to a larger overshoot size before pruning is repeated.

The pruning ratio \(r\) controls the minimum DP exposure required for an ordinary learned candidate to remain eligible. To preserve consistency with Pruned BPE~\cite{shao2026prunedbpe}, DH-BPE uses the same boundary-token reference for defining the threshold. During standard BPE training, when the ordinary vocabulary reaches \(N_{\text{visible}}-N_{\text{reserved}}\), let \(t_b\) denote the most recently created BPE token. Its BPE-training frequency is denoted by \(F(t_b)\). This is the merge frequency recorded during standard BPE training and is distinct from the DP exposure \(E_{\mathrm{DP}}(t)\) defined in Subsection~\ref{subsec:min-token-dp}.

For pruning ratio \(r\), the exposure threshold is

\begin{equation}
\tau_r
=
\left\lceil rF(t_b)\right\rceil.
\label{eq:exposure-threshold}
\end{equation}

Equation~\eqref{eq:exposure-threshold} therefore determines the threshold used by the eligibility criterion in Equation~\eqref{eq:eligibility-criterion} of Subsection~\ref{subsec:hierarchical-pruning}. The two quantities play different roles. 
\(F(t_b)\) provides a fixed frequency reference taken from the target-size point of standard BPE training, whereas \(E_{\mathrm{DP}}(t)\) measures a candidate's actual use under exact minimum-token segmentation of the overshoot vocabulary.

The overshoot factor \(f\) and pruning ratio \(r\) consequently control complementary aspects of DH-BPE: \(f\) determines how many learned candidates are available for selection, while \(r\) determines the minimum DP exposure required for ordinary candidates to qualify for model visibility.

\subsection{Overall Vocabulary Construction Procedure}
\label{subsec:overall-procedure}

Algorithm~\ref{alg:dh-bpe} summarizes the overall DH-BPE vocabulary construction procedure. For a target model-visible vocabulary size \(N_{\text{visible}}\), the required number of learned visible tokens \(N_{\text{required}}\) is determined by Equation~\eqref{eq:n-required}. Standard BPE is then trained to the overshoot size \(N_{\text{train}}\) defined by Equation~\eqref{eq:n-train}. During this training, when the ordinary vocabulary reaches \(N_{\text{visible}}-N_{\text{reserved}}\), DH-BPE records the BPE-training frequency \(F(t_b)\) of the boundary token \(t_b\). The pruning ratio \(r\) then determines the exposure threshold \(\tau_r\) according to Equation~\eqref{eq:exposure-threshold}.

DH-BPE then performs exact minimum-token segmentation of the training corpus using the overshoot vocabulary and records the DP exposure \(E_{\mathrm{DP}}(t)\) of each learned candidate. Candidates satisfying the exposure threshold are considered for the final model-visible vocabulary, while protected two-space tokens are retained independently of this threshold. Because BPE creates each merged token only after its two components already exist, the learned tokens are naturally ordered according to their dependency structure. 
DP exposure is used only to determine candidate eligibility; among eligible candidates, final selection follows this original BPE merge order rather than ranking candidates by exposure magnitude. DH-BPE temporarily retains the BPE prefix required by the latest selected visible token so that these dependencies remain valid during vocabulary construction. Tokens retained only for this purpose are not included in the vocabulary supplied to the final encoder.

The resulting model-visible vocabulary contains exactly \(N_{\text{visible}}\) tokens: the 256 byte-level tokens \(V_{\text{byte}}\), the selected learned tokens \(V_{\text{learn}}\), and the reserved tokens \(V_{\text{reserved}}\). Thus,
\(V_{\text{visible}} = V_{\text{byte}} \cup V_{\text{learn}} \cup V_{\text{reserved}}\). The construction-time internal tokens and other unselected candidates are not part of the final encoding vocabulary.

\begin{algorithm}[!ht]
\caption{DH-BPE Vocabulary Construction}
\label{alg:dh-bpe}
\scriptsize
\begin{algorithmic}[1]

\Require Training corpus \(C\);
target visible vocabulary size \(N_{\text{visible}}\);
reserved-token set \(V_{\text{reserved}}\);
overshoot factor \(f\);
pruning ratio \(r\)

\Ensure Final model-visible vocabulary \(V_{\text{visible}}\)

\State \(V_{\text{byte}} \gets\) the 256 single-byte tokens
\State \(N_{\text{byte}} \gets |V_{\text{byte}}| = 256\)
\State \(N_{\text{reserved}} \gets |V_{\text{reserved}}|\)

\State \(N_{\text{required}}
       \gets N_{\text{visible}}
       - N_{\text{reserved}}
       - N_{\text{byte}}\)

\State \(N_{\text{train}}
       \gets N_{\text{byte}}
       + \left\lceil fN_{\text{required}}\right\rceil\)

\State Train standard BPE in merge order

\State At ordinary vocabulary size
       \(N_{\text{visible}}-N_{\text{reserved}}\),
       let \(t_b\) be the most recently created token

\State Record its BPE-training frequency \(F(t_b)\)

\State Continue standard BPE training until the ordinary vocabulary
       reaches \(N_{\text{train}}\)

\State Add \(V_{\text{reserved}}\) to form the
       overshoot vocabulary \(V_{\text{cand}}\)

\State \(\tau_r \gets \left\lceil rF(t_b)\right\rceil\)

\State Segment \(C\) using exact minimum-token DP with \(V_{\text{cand}}\)

\State For each learned candidate \(t\), record its DP exposure
       \(E_{\mathrm{DP}}(t)\)

\State Let \(B=(t_1,t_2,\ldots,t_m)\) be the learned BPE candidates
       in original BPE merge order, so that each learned token appears 
       after its BPE children

\State \(P \gets
       \{t\in B :
       t \text{ begins with two consecutive spaces}\}\)

\State \(Q \gets
       (t_i\in B\setminus P :
       E_{\mathrm{DP}}(t_i)\ge\tau_r)\),
       retaining the order inherited from \(B\)

\If{\(|P|+|Q| < N_{\text{required}}\)}
    \State Increase the overshoot factor \(f\) and repeat the procedure
\EndIf

\State Let \(Q^\ast\) contain the first
       \(N_{\text{required}}-|P|\) candidates of the ordered sequence \(Q\)

\Statex \hspace{\algorithmicindent}
       \textit{DP exposure determines eligibility; eligible candidates are 
       selected in original BPE merge order rather than ranked by exposure magnitude}

\State \(V_{\text{learn}} \gets P\cup Q^\ast\)

\State Let \(t_j\) be the latest-created token in \(V_{\text{learn}}\)
       according to the BPE merge order

\State \(V_{\text{internal}} \gets
       \{t_i\in B : i\le j,\; t_i\notin V_{\text{learn}}\}\)

\Statex \hspace{\algorithmicindent}
       \textit{Retaining the BPE prefix through \(t_j\) conservatively
       preserves the dependencies of retained learned tokens}

\State Remap token IDs and BPE child IDs consistently so that byte tokens
       are followed by \(V_{\text{learn}}\) in BPE order and then by
       reserved tokens

\State \(V_{\text{visible}}
       \gets
       V_{\text{byte}}
       \cup V_{\text{learn}}
       \cup V_{\text{reserved}}\)

\State Discard \(V_{\text{internal}}\) for encoding

\State \Return \(V_{\text{visible}}\)

\end{algorithmic}
\end{algorithm}

\section{Experiments}
\label{sec:experiments}

\subsection{Training and Evaluation Data}
\label{subsec:training-evaluation-data}

\paragraph{Training corpora.}
The experiments reuse Corpus I and Corpus II from our previous Pruned BPE
work~\cite{shao2026prunedbpe}. Both corpora are used for tokenizer training and also serve as same-corpus or cross-corpus evaluation data, depending on the experimental configuration. Corpus I contains approximately 640~MB of UTF-8 text, consisting primarily of English and Chinese data together with smaller multilingual and code-oriented components. Corpus II contains approximately 1~GB and provides a larger mixture of English, Chinese, multilingual, and code-oriented text. Although the data distributions and topic mixtures of the two corpora are not identical, they are reasonably close in domain coverage: in particular, both contain Chinese Wikipedia text and English Reddit data~\cite{dewarim2017reddit}. The detailed data sources, sampling procedures, preprocessing, and corpus composition are described in our previous work~\cite{shao2026prunedbpe}.

\paragraph{Evaluation and parameter-tuning corpus.}
To study cross-corpus behavior under distribution shift, we additionally construct Corpus III, an approximately 900~MB corpus used for cross-corpus evaluation and empirical parameter tuning. Corpus III is never used as input to BPE training, DP-exposure computation, or the vocabulary-pruning procedure for any fixed tokenizer configuration. It contains five subsets with different degrees of similarity to the training data: approximately 170~MB of Chinese
medical-domain text; 40~MB of Chinese Weibo text drawn from the same general source and with a similar distribution to the social-media data in Corpus I, but without overlapping samples; 90~MB of Simplified and Traditional Chinese Wikipedia text drawn from the same source type as the Wikipedia data in Corpus I and Corpus II, but using non-overlapping pages covering different topics; 240~MB of English legal text sampled from the CourtListener opinions component of the Pile of Law dataset~\cite{henderson2022pileoflaw}; and
363~MB of English Reddit text~\cite{dewarim2017reddit}, drawn from source files not used for the Reddit data in Corpus I and Corpus II. Among these subsets, the legal corpus represents the strongest domain shift, whereas the Weibo, Wikipedia, and Reddit subsets provide non-overlapping evaluation data that remain closer in source or distribution to portions of the training corpora.

\paragraph{Pretokenization.}
All training and evaluation data use the same lightweight, Unicode-aware pretokenization procedure introduced in our previous work~\cite{shao2026prunedbpe}. The procedure preserves the original input exactly while separating obvious boundaries among word-like spans, numbers, punctuation, whitespace, scripts, and code-oriented structures. Runs of multiple spaces, tabs, newlines, and indentation are preserved rather than normalized. The same pretokenization is applied consistently during BPE training, DP exposure analysis, and final encoding; further implementation details are given in the previous paper~\cite{shao2026prunedbpe}.

\paragraph{Code and data availability.}
The Python implementation, corpus data, and trained tokenizer files used in this work will be made publicly available through the project repository~\cite{shao2026prunedbpegit}. The repository is intended to support reproduction of the reported results and tokenizer behavior.

\subsection{Compared Methods and Evaluation Methodology}
\label{subsec:evaluation-methodology}
We compare DH-BPE with Standard BPE, Pruned BPE, MinGram, and MinGram-PP. Standard BPE serves as the conventional baseline and uses the same training data and pretokenization procedure as DH-BPE. Pruned BPE~\cite{shao2026prunedbpe} is included to measure the improvement of DH-BPE over our earlier vocabulary-pruning approach. MinGram and MinGram-PP
are included as the closest compression-oriented alternatives. As discussed in Subsection~\ref{subsec:related-mingram}, both use BPE-derived candidate vocabularies and minimum-token information during vocabulary construction, but differ from DH-BPE in their final vocabulary-selection mechanisms.

For each experiment, methods are compared at the same final model-visible vocabulary size and, unless otherwise stated, are trained from the same corpus. The 256 byte-level tokens and the same reserved tokens are included within the reported vocabulary size for all configurations. Each method is given the candidate-vocabulary depth required by its own construction procedure; therefore, the overshoot factor is treated as a method-specific training parameter rather than requiring all methods to begin from an identically sized candidate vocabulary. For DH-BPE, the effects of the overshoot factor and pruning ratio are examined separately before selecting the configurations used in the main comparison. All methods use the same pretokenization procedure described in
Subsection~\ref{subsec:training-evaluation-data}.

Tokenizer efficiency is evaluated primarily by the total number of tokens required to encode an evaluation corpus. For a tokenizer \(T\) and evaluation corpus \(C_{\mathrm{eval}}\), we denote the encoded length by

\[
L(T,C_{\mathrm{eval}})
=
\sum_{x\in C_{\mathrm{eval}}}
|T(x)|,
\]

where \(T(x)\) is the token sequence produced for input \(x\). Lower encoded length indicates better compression at the same model-visible vocabulary size.

Standard BPE serves as the common baseline for the four alternative vocabulary-construction methods. Let \(T_{\mathrm{BPE}}\) denote the tokenizer constructed from the Standard BPE vocabulary and let \(T_M\) denote a tokenizer constructed by method \(M\), where
\(M \in \{\text{Pruned BPE},\text{MinGram},\text{MinGram-PP},\text{DH-BPE}\}\).

The relative encoded-length reduction of method \(M\) over Standard BPE is

\[
G_M
=
\frac{
L(T_{\mathrm{BPE}},C_{\mathrm{eval}})
-
L(T_M,C_{\mathrm{eval}})
}{
L(T_{\mathrm{BPE}},C_{\mathrm{eval}})
}
\times 100\%.
\]

A positive \(G_M\) therefore indicates that method \(M\) produces fewer tokens than Standard BPE. All such comparisons use the same final model-visible vocabulary size, training corpus, and evaluation corpus. To isolate the effect of vocabulary construction, all five methods use the same exact minimum-token DP encoder described in Subsection~\ref{subsec:min-token-dp} when encoding the evaluation corpus.

``Standard BPE'' therefore refers to the vocabulary produced by standard BPE training rather than to the conventional merge-rank BPE encoding procedure. Similarly, Pruned BPE is evaluated here using only its final model-visible \texttt{vocab.txt} with minimum-token DP encoding. This differs from the encoding procedure in our previous work
~\cite{shao2026prunedbpe}, which additionally used the internal-only \texttt{inter\_vocab.txt} and an encoding procedure derived from standard BPE. The present evaluation intentionally applies the same minimum-token encoder to all methods so that differences in encoded length reflect differences in the constructed model-visible vocabularies rather than differences in their original encoding procedures.

The Pruned BPE pruning ratio is tuned separately for each target vocabulary size under the shared minimum-token DP encoder. Our previous work evaluated \(r\in\{0.20,0.30,0.40\}\) using the native Pruned BPE encoder, which preserves the BPE merge structure and recursively expands internal-only tokens. In the present work, Pruned BPE is evaluated using only its
model-visible vocabulary under the shared minimum-token DP encoder. Re-evaluation under this encoding regime showed a modest benefit from higher pruning ratios, with \(r=0.70\) giving the lowest aggregate encoded length at 12K and \(r=0.65\) at 16K. We therefore use these respective settings in the main comparisons presented in 
Subsection~\ref{subsec:final-comparison}.

We evaluate both same-corpus and cross-corpus behavior. In a same-corpus evaluation, a tokenizer is evaluated on the corpus from which its vocabulary was constructed, measuring how effectively each method allocates vocabulary capacity to its training distribution. In a cross-corpus evaluation, the same trained tokenizer is applied without modification to a different corpus. Corpus I and Corpus II can therefore serve either as training or evaluation data depending on the experiment. Corpus III is not used for tokenizer training, DP-exposure computation, or vocabulary pruning, but is used both for cross-corpus evaluation and for empirical tuning of DH-BPE design and parameter settings. The individual subsets of Corpus III are also evaluated separately so that performance can be examined across domains with different degrees of similarity to the training data, including the substantially different English legal domain.

The DH-BPE parameter ranges were chosen to cover the empirically useful operating region while also including nearby settings that reveal performance degradation. We use \(f=1.07\) as the lower boundary of the overshoot-factor range because it is approximately the smallest overshoot tested that provides enough eligible candidates to fill the target vocabulary under a pruning ratio of \(r=0.70\) in most of our training configurations. Preliminary experiments further indicate that performance generally reaches its best region for \(r\) between \(0.70\) and \(0.80\), while degradation becomes clearly observable as the overshoot factor increases to approximately \(f=1.12\). These observations motivate the representative settings used in the two-space-token experiment in the following subsection, which combines \(f\in\{1.07,1.12\}\) with \(r\in\{0.70,0.80\}\). The subsequent parameter studies examine a broader surrounding range, using overshoot factors from \(1.07\) to \(1.15\) and pruning ratios from \(0.65\) to \(0.80\), to characterize these effects more completely.

\subsection{Effect of Two-Space Token Protection}
\label{subsec:two-space-experiment}
We briefly evaluate the two-space-token protection introduced in
Subsection~\ref{subsec:two-space-protection}. Here, \(H\) denotes ordinary DH-BPE pruning, while \(H\text{-}SP2\) denotes the variant that additionally protects learned tokens beginning with two consecutive ASCII spaces. Let \(L_H\) and \(L_{H\text{-}SP2}\) denote the encoded lengths produced by \(H\) and \(H\text{-}SP2\), respectively, on the evaluation corpus \(C_{\mathrm{eval}}\). Both variants are trained on Corpus I+II with an
18K target vocabulary and evaluated at \(f\in\{1.07,1.12\}\) and 
\(r\in\{0.70,0.80\}\). Positive values of \(L_H-L_{H\text{-}SP2}\) indicate fewer encoded tokens with two-space protection.

\begin{figure}[!ht]
\centering

\begin{tikzpicture}
\begin{groupplot}[
    group style={
        group size=2 by 2,
        horizontal sep=1.0cm,
        vertical sep=1.35cm
    },
    width=0.47\textwidth,
    height=0.29\textwidth,
    ybar,
    /pgf/bar width=5pt,
    symbolic x coords={
        Med,Weibo,Wiki,Reddit,Legal,CIII,CI,CII,C12
    },
    xtick=data,
    xticklabels={
        Med.,Weibo,Wiki,Reddit,Legal,CIII,CI,CII,CI+II
    },
    xticklabel style={
        rotate=45,
        anchor=east,
        font=\tiny
    },
    tick label style={font=\tiny},
    title style={font=\scriptsize},
    ylabel style={font=\scriptsize},
    ymajorgrids=true,
    grid style={dotted},
]

\nextgroupplot[
    title={(a) \(f=1.07,\ r=0.70\)},
    ylabel={\(L_H-L_{H\text{-}SP2}\) (\(10^3\) tokens)},
    ymin=-10,
    ymax=210
]
\addplot coordinates {
    (Med,0)
    (Weibo,0)
    (Wiki,0.014)
    (Reddit,-1.145)
    (Legal,195.491)
    (CIII,194.360)
    (CI,4.045)
    (CII,-0.403)
    (C12,3.642)
};

\nextgroupplot[
    title={(b) \(f=1.07,\ r=0.80\)},
    ymin=-50,
    ymax=1600
]
\addplot coordinates {
    (Med,-0.112)
    (Weibo,-0.413)
    (Wiki,-0.502)
    (Reddit,-0.065)
    (Legal,1515.731)
    (CIII,1514.639)
    (CI,64.991)
    (CII,14.397)
    (C12,79.388)
};

\nextgroupplot[
    title={(c) \(f=1.12,\ r=0.70\)},
    ylabel={\(L_H-L_{H\text{-}SP2}\) (\(10^3\) tokens)},
    ymin=-10,
    ymax=210
]
\addplot coordinates {
    (Med,-0.167)
    (Weibo,-0.133)
    (Wiki,-1.399)
    (Reddit,-0.195)
    (Legal,193.278)
    (CIII,191.384)
    (CI,4.038)
    (CII,-5.764)
    (C12,-1.726)
};

\nextgroupplot[
    title={(d) \(f=1.12,\ r=0.80\)},
    ymin=-50,
    ymax=1600
]
\addplot coordinates {
    (Med,-0.033)
    (Weibo,-0.014)
    (Wiki,-0.022)
    (Reddit,-2.619)
    (Legal,1516.038)
    (CIII,1513.350)
    (CI,66.398)
    (CII,13.561)
    (C12,79.959)
};

\end{groupplot}
\end{tikzpicture}

\caption{Effect of two-space token protection across four DH-BPE configurations}
\label{fig:sp2}
\end{figure}
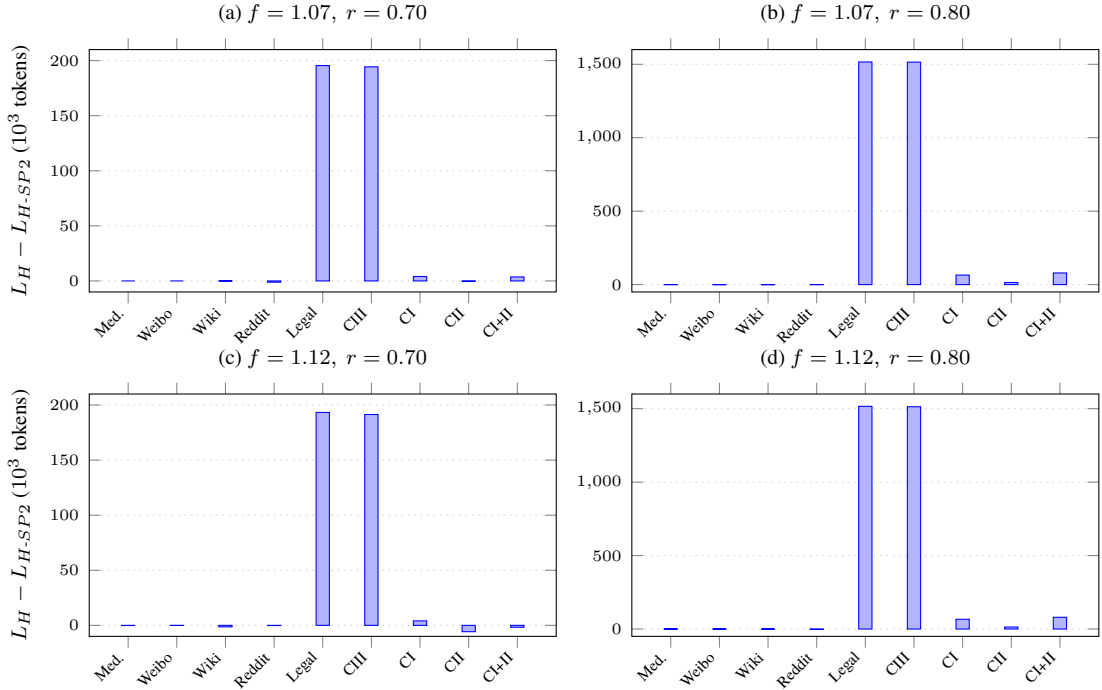

Figure~\ref{fig:sp2} shows that two-space protection has little effect on most evaluation subsets but substantially improves English Legal, particularly under the more 
aggressive \(r=0.80\) pruning setting, where the reduction exceeds 1.5 million tokens. The resulting improvement dominates the Corpus III total; smaller gains also appear on Corpus I+II at \(r=0.80\). This indicates that exposure-based pruning can occasionally remove repeated-whitespace tokens that remain useful in formatting-sensitive text. Because the safeguard prevents this failure mode while otherwise producing only small changes, we use two-space protection in all subsequent DH-BPE experiments.

\subsection{Effect of the Overshoot Factor}
\label{subsec:overshoot-factor}
We next examine how the overshoot factor \(f\) affects the final DH-BPE vocabulary. As defined in Subsection~\ref{subsec:overshoot-pruning-ratio}, \(f\) controls how far standard BPE training proceeds beyond the target model-visible vocabulary size before DP exposure analysis and hierarchical pruning are applied. The primary experiment uses DH-BPE tokenizers \(T_{f,r}\) trained on the combined Corpus I+II data with a target model-visible vocabulary size of 18K, where \(f\) denotes the overshoot factor and \(r\) the pruning ratio. We evaluate five overshoot factors, \(f\in\{1.07,1.08,1.10,1.12,1.15\}\), at pruning ratios \(r\in\{0.70,0.75,0.80\}\). Results are reported separately on the combined Corpus I+II data, representing same-corpus evaluation, and on Corpus III, representing cross-corpus evaluation. To make the effect of \(f\) visible despite the large absolute encoded lengths, each curve is normalized by the best result obtained within the same pruning ratio. Specifically, for evaluation corpus \(C_{\mathrm{eval}}\), we plot

\[
\Delta L(f,r;C_{\mathrm{eval}})
=
L(T_{f,r},C_{\mathrm{eval}})
-
\min_{f'\in\mathcal{O}}
L(T_{f',r},C_{\mathrm{eval}}),
\]

where \(\mathcal{O}\) is the set of overshoot factors evaluated in the corresponding experiment. Thus, \(\Delta L=0\) identifies the best tested overshoot factor for a given pruning ratio, while larger values indicate additional encoded tokens.

\begin{figure}[!ht]
\centering

\begin{tikzpicture}
\begin{groupplot}[
    group style={
        group size=2 by 1,
        horizontal sep=1.2cm
    },
    width=0.47\textwidth,
    height=0.34\textwidth,
    xlabel={Overshoot factor \(f\)},
    xmin=1.065,
    xmax=1.155,
    ymin=0,
    ymax=200,
    xtick={1.07,1.08,1.10,1.12,1.15},
    xticklabels={1.07,1.08,1.10,1.12,1.15},
    tick label style={font=\scriptsize},
    label style={font=\scriptsize},
    title style={font=\scriptsize},
    legend style={
        font=\scriptsize,
        at={(0.03,0.97)},
        anchor=north west
    },
    ymajorgrids=true,
    grid style={dotted},
]

\nextgroupplot[
    title={(a) Corpus I+II},
    ylabel={Excess encoded tokens (\(10^3\))},
]

\addplot+[mark=*] coordinates {
    (1.07,0)
    (1.08,41.010)
    (1.10,59.299)
    (1.12,110.166)
    (1.15,185.870)
};
\addlegendentry{\(r=0.70\)}

\addplot+[mark=square*] coordinates {
    (1.07,0)
    (1.08,42.600)
    (1.10,67.011)
    (1.12,98.725)
    (1.15,145.490)
};
\addlegendentry{\(r=0.75\)}

\addplot+[mark=triangle*] coordinates {
    (1.07,0)
    (1.08,39.137)
    (1.10,66.011)
    (1.12,86.682)
    (1.15,164.754)
};
\addlegendentry{\(r=0.80\)}

\nextgroupplot[
    title={(b) Corpus III},
]

\addplot+[mark=*] coordinates {
    (1.07,23.860)
    (1.08,18.984)
    (1.10,0)
    (1.12,5.534)
    (1.15,47.087)
};
\addlegendentry{\(r=0.70\)}

\addplot+[mark=square*] coordinates {
    (1.07,11.905)
    (1.08,0)
    (1.10,7.528)
    (1.12,23.627)
    (1.15,64.825)
};
\addlegendentry{\(r=0.75\)}

\addplot+[mark=triangle*] coordinates {
    (1.07,0)
    (1.08,11.057)
    (1.10,42.087)
    (1.12,35.588)
    (1.15,69.455)
};
\addlegendentry{\(r=0.80\)}

\end{groupplot}
\end{tikzpicture}

\caption{Effect of the overshoot factor for DH-BPE trained on Corpus I+II
at an 18K target vocabulary size}
\label{fig:overshoot-18k}
\end{figure}
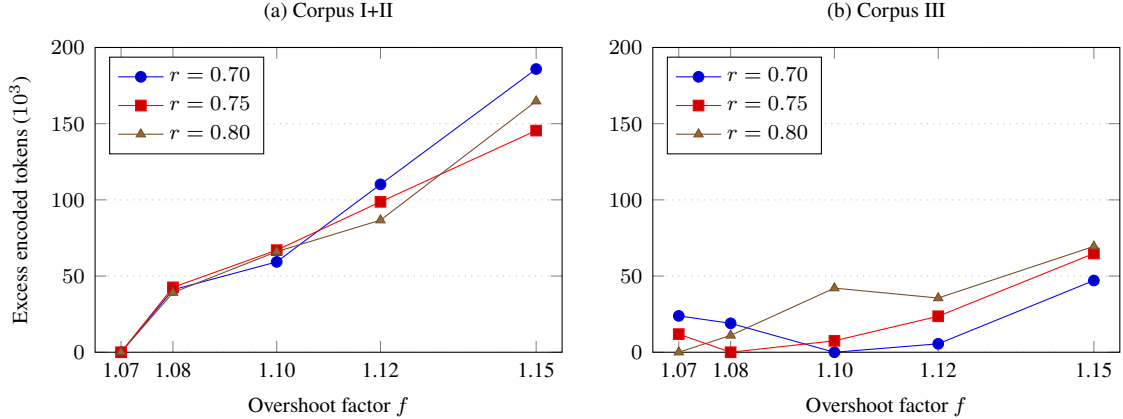

Figure~\ref{fig:overshoot-18k} shows a clear distinction between same-corpus
(left panel) and cross-corpus (right panel) behavior. Each curve shows the excess
encoded length relative to the best tested \(f\) for the same pruning ratio.

On Corpus I+II, \(f=1.07\) produces the shortest encoding at all three pruning
ratios, and the encoded length generally increases as the overshoot factor
grows. At \(f=1.15\), the penalties relative to the best result increase
sharply at all three tested pruning ratios, indicating that the deeper candidate
pool does not improve same-corpus compression in these configurations.

The cross-corpus results are less monotonic. On Corpus III, as \(r\) increases
from \(0.70\) to \(0.75\) and \(0.80\), the best tested overshoot factor \(f\)
decreases from \(1.10\) to \(1.08\) and \(1.07\), respectively. This indicates
that a modest increase in candidate depth can improve cross-corpus compression
in some settings, but a substantially larger overshoot can instead degrade
performance. In particular, \(f=1.15\) is the worst of the tested factors for
all three pruning ratios, with degradation already evident by \(f=1.12\).

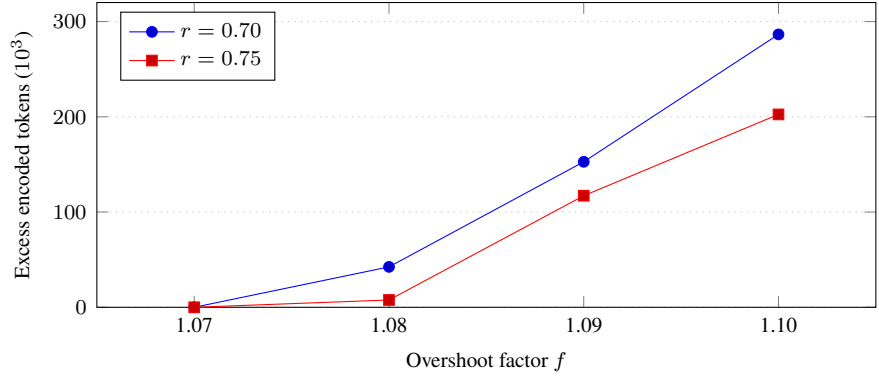
\begin{figure}[!ht]
\centering

\begin{tikzpicture}
\begin{axis}[
    width=0.72\textwidth,
    height=0.34\textwidth,
    xlabel={Overshoot factor \(f\)},
    ylabel={Excess encoded tokens (\(10^3\))},
    xmin=1.065,
    xmax=1.105,
    ymin=0,
    ymax=320,
    xtick={1.07,1.08,1.09,1.10},
    xticklabels={1.07,1.08,1.09,1.10},
    tick label style={font=\scriptsize},
    label style={font=\scriptsize},
    legend style={
        font=\scriptsize,
        at={(0.03,0.97)},
        anchor=north west
    },
    ymajorgrids=true,
    grid style={dotted},
]

\addplot+[mark=*] coordinates {
    (1.07,0)
    (1.08,42.416)
    (1.09,152.755)
    (1.10,286.523)
};
\addlegendentry{\(r=0.70\)}

\addplot+[mark=square*] coordinates {
    (1.07,0)
    (1.08,7.653)
    (1.09,117.130)
    (1.10,202.527)
};
\addlegendentry{\(r=0.75\)}

\end{axis}
\end{tikzpicture}

\caption{Effect of the overshoot factor for DH-BPE trained on Corpus II at a
12K target vocabulary size}
\label{fig:overshoot-12k}
\end{figure}

To determine whether this behavior is specific to the 18K Corpus I+II configuration, we additionally examine a smaller target vocabulary trained on Corpus II. 

Figure~\ref{fig:overshoot-12k} reports results for a 12K target vocabulary using
\(f\in\{1.07,1.08,1.09,1.10\}\) at \(r\in\{0.70,0.75\}\). For each configuration, encoded lengths are aggregated over evaluations on Corpus II, Corpus I, and Corpus III, and the resulting totals are shown as excess encoded tokens relative to the best tested \(f\) within the same pruning ratio. This experiment exhibits the same overall preference for a small overshoot factor. For \(r\in\{0.70,0.75\}\), \(f=1.07\) gives the lowest aggregate encoded
length. This result provides evidence that the low-overshoot preference observed previously is not limited to a single training corpus or vocabulary size.

A practical constraint, however, is that an overshoot factor that is too small may not provide enough eligible candidates to fill the target vocabulary under a higher pruning ratio. For example, in an additional 16K experiment, \(f=1.07\) does not provide enough eligible candidates at \(r=0.75\) to fill the target vocabulary. Taken together, the experiments suggest using the smallest overshoot factor that provides a sufficient eligible candidate pool, typically near \(f=1.07\)--\(1.08\) for the configurations examined here.

\subsection{Effect of the Pruning Ratio}
\label{subsec:pruning-ratio}
We next examine the effect of the pruning ratio \(r\), which determines the DP-exposure threshold through Equation~\eqref{eq:exposure-threshold}. The experiment uses DH-BPE tokenizers \(T_{f,r}\) trained on the combined Corpus I+II training data with a target model-visible vocabulary size of 18K. For \(f\in\{1.07,1.08,1.10\}\), we evaluate pruning ratios
\(r\in\{0.65,0.70,0.75,0.80\}\). For the larger overshoot factors previously found to be less favorable, \(f\in\{1.12,1.15\}\), we evaluate only \(r\in\{0.70,0.75,0.80\}\) to reduce the number of experimental configurations. 

As in the overshoot-factor analysis, we report same-corpus and cross-corpus behavior separately. To make the effect of \(r\) directly visible, each curve corresponds to a fixed overshoot factor and is normalized by the best pruning ratio evaluated for that factor.

\pgfplotsset{
    f107/.style={
        draw=blue,
        mark=*,
        mark options={draw=blue, fill=blue}
    },
    f108/.style={
        draw=red,
        mark=square*,
        mark options={draw=red, fill=red}
    },
    f110/.style={
        draw=green!60!black,
        mark=triangle*,
        mark options={draw=green!60!black, fill=green!60!black}
    },
    f112/.style={
        draw=orange!80!black,
        mark=diamond*,
        mark options={draw=orange!80!black, fill=orange!80!black}
    },
    f115/.style={
        draw=purple,
        mark=pentagon*,
        mark options={draw=purple, fill=purple}
    }
}

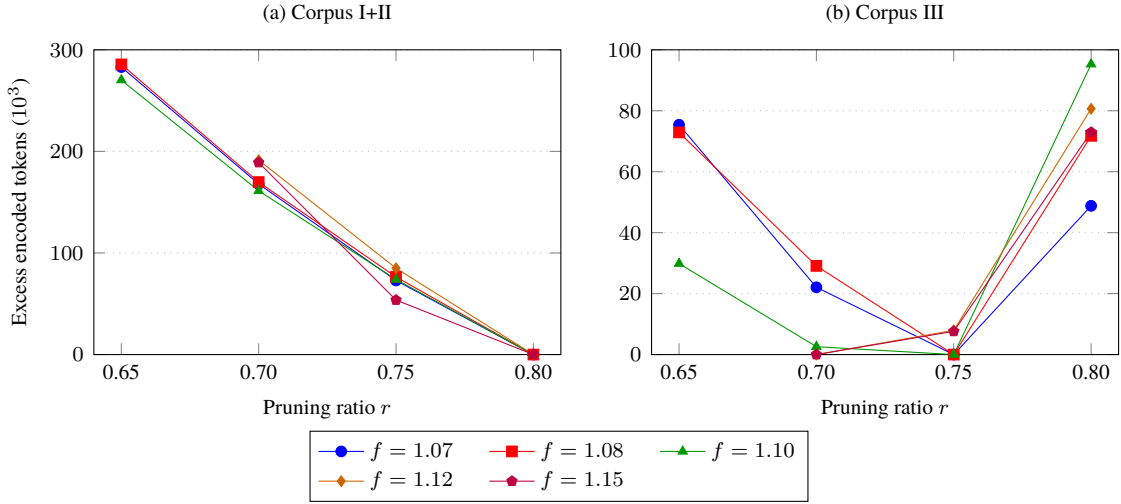
\begin{figure}[!ht]
\centering

\begin{tikzpicture}
\begin{groupplot}[
    group style={
        group size=2 by 1,
        horizontal sep=1.2cm
    },
    width=0.47\textwidth,
    height=0.34\textwidth,
    xlabel={Pruning ratio \(r\)},
    xmin=0.64,
    xmax=0.81,
    xtick={0.65,0.70,0.75,0.80},
    xticklabels={0.65,0.70,0.75,0.80},
    tick label style={font=\scriptsize},
    label style={font=\scriptsize},
    title style={font=\scriptsize},
    ymajorgrids=true,
    grid style={dotted},
]

\nextgroupplot[
    title={(a) Corpus I+II},
    ylabel={Excess encoded tokens (\(10^3\))},
    ymin=0,
    ymax=300,
    legend to name=pruneratiolegend,
    legend columns=3,
    legend style={
        font=\scriptsize,
        /tikz/every even column/.append style={column sep=0.4cm}
    }
]

\addplot[f107] coordinates {
    (0.65,283.036)
    (0.70,167.900)
    (0.75,72.953)
    (0.80,0)
};
\addlegendentry{\(f=1.07\)}

\addplot[f108] coordinates {
    (0.65,285.652)
    (0.70,169.773)
    (0.75,76.416)
    (0.80,0)
};
\addlegendentry{\(f=1.08\)}

\addplot[f110] coordinates {
    (0.65,270.205)
    (0.70,161.188)
    (0.75,73.953)
    (0.80,0)
};
\addlegendentry{\(f=1.10\)}

\addplot[f112] coordinates {
    (0.70,191.384)
    (0.75,84.996)
    (0.80,0)
};
\addlegendentry{\(f=1.12\)}

\addplot[f115] coordinates {
    (0.70,189.016)
    (0.75,53.689)
    (0.80,0)
};
\addlegendentry{\(f=1.15\)}

\nextgroupplot[
    title={(b) Corpus III},
    ymin=0,
    ymax=100
]

\addplot[f107] coordinates {
    (0.65,75.412)
    (0.70,22.075)
    (0.75,0)
    (0.80,48.808)
};

\addplot[f108] coordinates {
    (0.65,72.891)
    (0.70,29.104)
    (0.75,0)
    (0.80,71.770)
};

\addplot[f110] coordinates {
    (0.65,29.862)
    (0.70,2.592)
    (0.75,0)
    (0.80,95.272)
};

\addplot[f112] coordinates {
    (0.70,0)
    (0.75,7.973)
    (0.80,80.647)
};

\addplot[f115] coordinates {
    (0.70,0)
    (0.75,7.618)
    (0.80,72.961)
};

\end{groupplot}
\end{tikzpicture}

\ref{pruneratiolegend}

\caption{Effect of the pruning ratio for DH-BPE trained on Corpus I+II at an
18K target vocabulary size}
\label{fig:pruning-ratio}
\end{figure}

Figure~\ref{fig:pruning-ratio} reveals contrasting same-corpus (left panel) and cross-corpus (right panel) behavior. Each curve shows the excess encoded length relative to the best tested \(r\) for the same overshoot factor. The two panels use different vertical scales because the pruning-ratio effect is substantially larger on the same-corpus evaluation.

On Corpus I+II, increasing \(r\) consistently improves compression, and \(r=0.80\) is best for every evaluated overshoot factor. On Corpus III, however, the best configurations shift toward moderate pruning: \(r=0.75\) is best for \(f=1.07\)--\(1.10\), while \(r=0.70\) is best for \(f=1.12\) and \(f=1.15\). In every case, \(r=0.80\) is worse than the
cross-corpus optimum.

The results therefore suggest a trade-off between stronger specialization to the
training distribution and performance under distribution shift. Accordingly, we 
use \(r=0.70\)--\(0.75\) as the balanced operating range in the subsequent cross-method comparisons.

\subsection{Comparison with Standard BPE, Pruned BPE, MinGram, and MinGram-PP}
\label{subsec:final-comparison}

We now compare DH-BPE with Standard BPE, our previous Pruned BPE method, MinGram, and MinGram-PP. The primary comparison uses tokenizers trained on Corpus II with a target model-visible vocabulary size of 12K and evaluates each vocabulary on Corpus II, Corpus I, and Corpus III. This configuration provides the most complete comparison because MinGram-PP was evaluated with overshoot factors from \(f=2.0\) through \(f=5.0\). To avoid presenting multiple DH-BPE variants in the final comparison, we report \(f=1.07,\ r=0.70\), the DH-BPE configuration with the lowest aggregate encoded length across Corpus II, Corpus I, and Corpus III among the evaluated \(r=0.70\)--\(0.75\) configurations at this target size. Relative improvements \(G_M\) are calculated against Standard BPE as defined in Subsection~\ref{subsec:evaluation-methodology}.

\begin{table}[!ht]
\centering
\scriptsize
\setlength{\tabcolsep}{4pt}
\renewcommand{\arraystretch}{1.15}

\begin{tabular}{@{}llrrrrr@{}}
\toprule
\textbf{Method} &
\textbf{Configuration} &
\textbf{Corpus II} &
\textbf{Corpus I} &
\textbf{Corpus III} &
\textbf{Overall} &
\(\mathbf{G_M}\) \textbf{(\%)} \\
\midrule

Standard BPE &
-- &
307,026,651 &
190,730,625 &
276,009,112 &
773,766,388 &
0.000 \\

\midrule

Pruned BPE &
\(r=0.70\) &
305,341,252 &
190,006,412 &
274,706,279 &
770,053,943 &
0.480 \\

\midrule

MinGram &
\(f=1.15\) &
304,671,017 &
189,891,430 &
274,850,074 &
769,412,521 &
0.563 \\

\midrule

\multirow{4}{*}{MinGram-PP}
& \(f=2.0\) & 302,442,192 & 188,991,924 & 275,064,145 & 766,498,261 & 0.939 \\

& \(f=3.0\) & 302,224,084 & 189,055,077 & 275,376,899 & 766,656,060 & 0.919 \\

& \(f=4.0\) & 302,148,125 & 189,082,771 & 274,071,348 & 765,302,244 & 1.094 \\

& \(f=5.0\) & 302,177,315 & 189,024,302 & 274,000,931 & \textbf{765,202,548} & \textbf{1.107} \\

\midrule

\textbf{DH-BPE} &
\(f=1.07,\ r=0.70\) &
304,785,208 &
189,708,876 &
274,321,323 &
768,815,407 &
0.640 \\

\bottomrule
\end{tabular}

\caption{Comparison of methods trained on Corpus II at a 12K target vocabulary size}
\label{tab:final-comp-12k}
\end{table}

Table~\ref{tab:final-comp-12k} shows that all four alternative vocabulary-construction methods improve over the Standard BPE vocabulary. Pruned BPE reduces the aggregate encoded length by \(0.480\%\), while MinGram achieves a reduction of \(0.563\%\). DH-BPE improves this further to \(0.640\%\), outperforming both our previous Pruned BPE method and
MinGram at the same final model-visible vocabulary size.

The cross-corpus results on Corpus III reveal a different pattern. With the relatively small overshoot factor \(f=1.07\), DH-BPE outperforms MinGram-PP at \(f=2.0\) and \(f=3.0\). MinGram-PP reverses this ordering only at the larger overshoot factors \(f=4.0\) and \(f=5.0\). Thus, while MinGram-PP achieves the strongest overall compression in this experiment, its Corpus III advantage over DH-BPE appears only with a substantially larger candidate vocabulary among the configurations evaluated here.

Across all three evaluation corpora, MinGram-PP obtains the strongest aggregate compression, with its best result at \(f=5.0\), followed closely by \(f=4.0\). The differences on Corpus III, however, are not uniform across domains, motivating the subset-level comparison below.

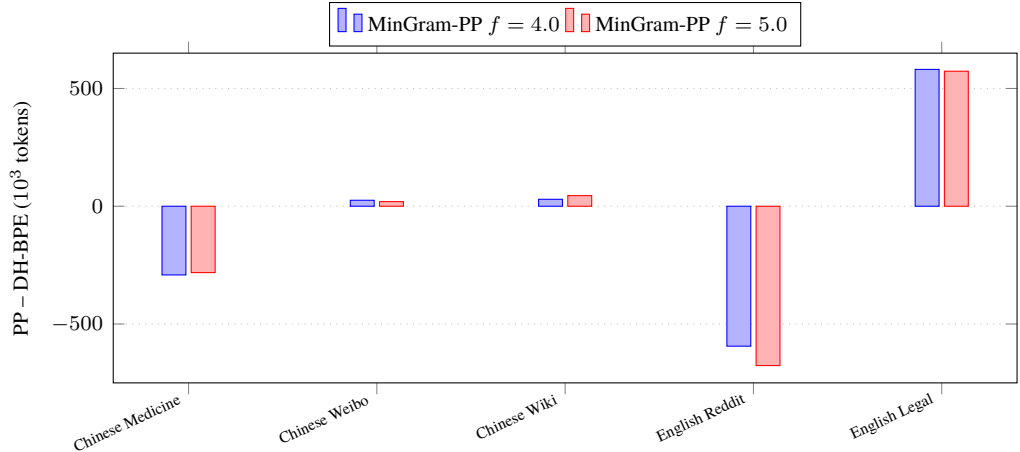
\begin{figure}[!ht]
\centering

\begin{tikzpicture}
\begin{axis}[
    width=0.82\textwidth,
    height=0.36\textwidth,
    ybar,
    /pgf/bar width=9pt,
    symbolic x coords={
        Medicine,Weibo,Wiki,Reddit,Legal
    },
    xtick=data,
    xticklabels={
        Chinese Medicine,
        Chinese Weibo,
        Chinese Wiki,
        English Reddit,
        English Legal
    },
    xticklabel style={
        rotate=25,
        anchor=east,
        font=\tiny
    },
    ylabel={PP -- DH-BPE (\(10^3\) tokens)},
    tick label style={font=\scriptsize},
    label style={font=\scriptsize},
    legend style={
        font=\scriptsize,
        at={(0.5,1.02)},
        anchor=south,
        legend columns=2
    },
    ymin=-750,
    ymax=650,
    ymajorgrids=true,
    grid style={dotted},
]

\addplot coordinates {
    (Medicine,-291.711)
    (Weibo,25.389)
    (Wiki,29.484)
    (Reddit,-594.142)
    (Legal,581.005)
};
\addlegendentry{MinGram-PP \(f=4.0\)}

\addplot coordinates {
    (Medicine,-281.788)
    (Weibo,19.560)
    (Wiki,45.073)
    (Reddit,-676.493)
    (Legal,573.256)
};
\addlegendentry{MinGram-PP \(f=5.0\)}

\end{axis}
\end{tikzpicture}

\caption{Encoded-length differences on Corpus III subsets for Corpus II-trained 12K vocabularies}
\label{fig:pp-dh-corpus3}
\end{figure}

\paragraph{Cross-domain behavior on Corpus III.}
The 12K results show that DH-BPE outperforms MinGram-PP at \(f=2.0\) and \(f=3.0\) on aggregate Corpus III, while MinGram-PP reverses this ordering at \(f=4.0\) and \(f=5.0\). To examine where the advantage of these larger MinGram-PP configurations arises, Figure~\ref{fig:pp-dh-corpus3} compares DH-BPE (\(f=1.07,\ r=0.70\)) with MinGram-PP at \(f=4.0\) and \(f=5.0\) on the five Corpus III subsets. Positive values indicate that DH-BPE produces fewer tokens, whereas negative values indicate that MinGram-PP produces fewer tokens.

Figure~\ref{fig:pp-dh-corpus3} shows that the aggregate advantage of MinGram-PP is not uniform across domains. MinGram-PP performs substantially better on Chinese Medicine and English Reddit. In contrast, DH-BPE performs better on Chinese Weibo, Chinese Wikipedia, and English Legal. The largest DH-BPE advantage occurs on English Legal, where it produces 581,005 fewer tokens than MinGram-PP at \(f=4.0\) and 573,256 fewer at \(f=5.0\). The Weibo and Wikipedia advantages are substantially smaller but remain positive for both MinGram-PP configurations.

This domain-dependent behavior is particularly notable because English Legal represents the strongest distribution shift among the Corpus III subsets, whereas English Reddit is drawn from the same source type as data represented in the training corpora. The results therefore show that MinGram-PP's aggregate compression advantage does not translate uniformly across domains. The qualitative vocabulary analysis in Subsection~\ref{subsec:qualitative-tokens} examines the retained tokens of these methods to provide further insight into this difference.

\paragraph{Comparison at a 16K target vocabulary.}
We additionally repeat the broad comparison with tokenizers trained on Corpus II at a larger target vocabulary size of 16K. To keep this supporting comparison compact, one representative configuration is shown for each method. For DH-BPE, \(f=1.08,\ r=0.75\) gives the lowest aggregate encoded length across Corpus II, Corpus I, and Corpus III among the evaluated configurations in the balanced \(r=0.70\)--\(0.75\) range. MinGram-PP is reported at the two available overshoot factors, \(f=2.0\) and \(f=3.0\), at this target size. Because larger MinGram-PP overshoot factors were not evaluated at 16K, these results should not be interpreted as having established its optimal configuration at this vocabulary size.

\begin{table}[!ht]
\centering
\scriptsize
\setlength{\tabcolsep}{4pt}
\renewcommand{\arraystretch}{1.15}

\begin{tabular}{@{}llrrrrr@{}}
\toprule
\textbf{Method} &
\textbf{Configuration} &
\textbf{Corpus II} &
\textbf{Corpus I} &
\textbf{Corpus III} &
\textbf{Overall} &
\(\mathbf{G_M}\) \textbf{(\%)} \\
\midrule

Standard BPE &
-- &
292,257,884 &
182,736,861 &
263,421,152 &
738,415,897 &
0.000 \\

\midrule

Pruned BPE &
\(r=0.65\) &
290,775,983 &
181,962,160 & 
262,531,829 &
735,269,972 &
0.426 \\

\midrule

MinGram &
\(f=1.15\) &
290,172,998 &
181,990,067 &
263,154,525 &
735,317,590 &
0.420 \\

\midrule

\multirow{2}{*}{MinGram-PP}

& \(f=2.0\) & 288,217,401 & 181,301,272 & 263,708,429 & 733,227,102 & 0.703 \\

& \(f=3.0\) & 287,975,182 & 181,301,360 & 262,688,176 & \textbf{731,964,718} &
\textbf{0.874} \\

\midrule

\textbf{DH-BPE} &
\(f=1.08,\ r=0.75\) &
290,193,410 &
181,843,497 &
\textbf{262,355,943} &
734,392,850 &
0.545 \\

\bottomrule
\end{tabular}

\caption{Supporting comparison using Corpus II training at a 16K target vocabulary size}
\label{tab:final-comp-16k}
\end{table}

At 16K, DH-BPE again improves aggregate compression over both Pruned BPE and MinGram, while MinGram-PP achieves the strongest aggregate compression at the two tested overshoot factors.

On Corpus III, however, DH-BPE outperforms MinGram-PP at both \(f=2.0\) and \(f=3.0\), showing that the aggregate advantage of MinGram-PP does not extend uniformly to cross-corpus evaluation. Because the 16K MinGram-PP experiments were performed only through \(f=3.0\), we do not claim that DH-BPE outperforms the best possible MinGram-PP configuration at this vocabulary size. The corresponding 12K results show that larger MinGram-PP overshoot factors can reverse this ordering, so the 16K comparison should be interpreted as evidence that DH-BPE remains competitive in cross-corpus compression while using a substantially smaller candidate overshoot.

\paragraph{Additional 18K comparison.}
Finally, the Corpus I+II-trained 18K experiment provides an additional comparison between DH-BPE and MinGram under a different training distribution and target vocabulary size. Using DH-BPE with \(f=1.07,\ r=0.75\), MinGram produces the shorter encoding on the same-corpus
Corpus I+II evaluation: 460,138,407 tokens compared with 460,297,165 for DH-BPE, an advantage of 158,758 tokens. The ordering reverses on Corpus III, where DH-BPE requires 254,640,541 tokens compared with 254,869,851 for MinGram, giving DH-BPE an advantage of 229,310 tokens. Across the combined Corpus I+II and Corpus III evaluations, DH-BPE therefore produces 70,552 fewer tokens overall. This additional experiment reinforces the observation that the relative compression performance of the learned vocabularies can depend on the
relationship between the training and evaluation distributions, and that same-corpus compression alone does not fully characterize cross-corpus behavior.

In summary, these comparisons show three consistent patterns. First, at both 12K and 16K, 
DH-BPE improves aggregate compression over Standard BPE, Pruned BPE, and MinGram, 
demonstrating that the resulting combination of DP-guided exposure and BPE-hierarchical selection provides stronger aggregate compression than these baselines in the evaluated configurations. Second, MinGram-PP achieves the strongest aggregate compression when sufficiently large overshoot factors are used, but this ordering does not transfer uniformly across evaluation domains: DH-BPE remains competitive on Corpus III and outperforms MinGram-PP on several of its constituent domains despite using a much smaller overshoot factor. Third, the 18K Corpus I+II experiment further shows that relative vocabulary quality can change under distribution shift, with MinGram performing better on
the training-distribution evaluation while DH-BPE performs better on Corpus III.
These results suggest that aggregate in-distribution compression and cross-corpus
robustness capture complementary aspects of vocabulary quality. The qualitative analysis in Subsection~\ref{subsec:qualitative-tokens} examines the retained tokens directly to better understand the vocabulary-selection behavior underlying these differences.

\subsection{Qualitative Analysis of Retained Tokens}
\label{subsec:qualitative-tokens}
The preceding results evaluate vocabulary quality through corpus-level token counts. To complement these aggregate measurements, we examine the tokens that distinguish DH-BPE from the other pruning methods. The purpose of this analysis is not to assign an intrinsic quality to individual tokens, but to characterize the different vocabulary-selection behaviors induced by the methods. All vocabularies considered here are trained on Corpus~II with a target vocabulary size of 12K. We use DH-BPE with \(f=1.07\) and \(r=0.70\) throughout, and compare it against Pruned BPE with a \(70\%\) pruning ratio, MinGram with \(f=1.15\), and MinGram-PP with \(f=5.0\). For each pair, we inspect tokens that occur exclusively in one of the two final vocabularies. Leading whitespace is omitted from token examples for readability. Because the methods use different candidate-pool sizes and pruning procedures, these comparisons are descriptive rather than controlled ablations of a single design choice.

\paragraph{Comparison with Pruned BPE.}
DH-BPE and Pruned BPE produce highly similar vocabularies: 11,755 of the 12,000 tokens (\(97.96\%\)) are shared, leaving only 245 tokens unique to each method. Despite this high overlap, the differing tokens reveal a clear pattern. Many Pruned-BPE-only tokens are incomplete English stems or continuations, such as \texttt{Ariz}, \texttt{Tay}, \texttt{achie}, \texttt{appreci}, \texttt{challeng}, \texttt{convers}, and \texttt{respons}. In contrast, the DH-BPE-only set contains a number of more complete lexical units, including \texttt{Arizona}, \texttt{Alabama}, \texttt{Taylor}, \texttt{YouTube}, \texttt{birthday}, \texttt{developing}, \texttt{manufacture}, and \texttt{schedule}. Several differences directly illustrate this extension from a shorter component to a more complete token: Pruned BPE retains \texttt{Ariz}, \texttt{Tay}, and \texttt{Tube}, whereas DH-BPE retains \texttt{Arizona}, \texttt{Taylor}, and \texttt{YouTube}, respectively.

A similar effect appears in Chinese. Pruned-BPE-only tokens include shorter components such as \cjktok{流动}, \cjktok{莱坞}, and \cjktok{睛}, whereas DH-BPE uniquely retains \cjktok{流动性}, \cjktok{好莱坞}, and \cjktok{眼睛}, together with multi-character units such as \cjktok{一方面}, \cjktok{取得了}, \cjktok{真正的}, and \cjktok{相当于}. Thus, the modest BPE overshoot used by DH-BPE can expose useful later merges beyond the original target boundary, while DP-guided selection can allocate some of the final vocabulary slots to these extended units. Importantly, this occurs without substantially restructuring the vocabulary: fewer than \(2.1\%\) of the final entries differ between the two methods.

\paragraph{Comparison with MinGram.}
The difference from MinGram is somewhat larger but remains localized. The two vocabularies share 11,572 tokens (\(96.43\%\)), with 428 tokens unique to each. Here, however, the qualitative direction differs from the Pruned BPE comparison. MinGram uniquely retains many complete lexical items, including \texttt{behavior}, \texttt{exercise}, \texttt{function}, \texttt{influence}, \texttt{professor}, \texttt{religion}, and \texttt{ridiculous}. DH-BPE often retains a shorter reusable component instead; representative pairs include MinGram's \texttt{behavior}, \texttt{exercise}, \texttt{professor}, \texttt{ridiculous}, and \texttt{stretch}, compared with DH-BPE's \texttt{behav}, \texttt{exerc}, \texttt{profess}, \texttt{ridicul}, and \texttt{stret}. The same pattern is visible in Chinese: MinGram uniquely contains \cjktok{分辨率}, \cjktok{性价比}, and \cjktok{有利于}, while DH-BPE retains the shorter components \cjktok{分辨}, \cjktok{性价}, and \cjktok{有利}.

This contrast is consistent with the different final selection strategies. MinGram's minimum-token segmentation followed by flat score-based pruning can directly favor longer lexicalized units that occur on the selected minimum-token paths. DH-BPE instead retains candidates under a BPE-structured, DP-exposure-based criterion, and therefore may preserve an intermediate subword when that component remains useful across multiple larger constructions. Consequently, DH-BPE should not be interpreted as simply preferring longer tokens. Relative to Pruned BPE, its small overshoot can replace some incomplete target-boundary fragments with later, more complete merges; relative to MinGram, however, it can favor more compositional units over fully lexicalized alternatives.

\paragraph{Comparison with MinGram-PP.}
The largest qualitative difference occurs with MinGram-PP at \(f=5.0\). Only 10,070 tokens (\(83.92\%\)) are shared, and 1,930 entries are unique to each vocabulary. MinGram-PP uniquely retains a large number of complete words and named entities, including \texttt{Administration}, \texttt{Afghanistan}, \texttt{Charlottesville}, \texttt{Microsoft}, \texttt{Pennsylvania}, \texttt{Venezuela}, and many other geographic, political, sports, and organizational names. In contrast, the DH-BPE-only vocabulary contains substantially more reusable stems and continuations. Several related examples make the distinction particularly clear: MinGram-PP retains \texttt{communication}, \texttt{contribution}, \texttt{controversial}/\texttt{controversy}, \texttt{determine}, \texttt{manufacturing}, and \texttt{prosecution}/\texttt{prosecutor}, whereas DH-BPE uniquely retains the corresponding shorter forms \texttt{communic}, \texttt{contribut}, \texttt{controvers}, \texttt{determin}, \texttt{manufact}, and \texttt{prosecut}. These shorter units can participate in multiple surface forms rather than dedicating separate vocabulary entries to individual completed words.

The contrast is even more pronounced for Chinese. MinGram-PP retains highly lexicalized expressions such as \cjktok{中共中央政治局}, \cjktok{偏股型基金}, and \cjktok{以上信息仅供参考}, as well as unusually long corpus-specific strings such as \cjktok{仅为分析人士对一只股票的个人观点和看法}. Some retained entries extend to sentence-like boilerplate sequences, indicating that the much larger \(f=5.0\) candidate vocabulary makes very late BPE merges available for subsequent PathPiece-style pruning. DH-BPE instead retains shorter components such as \cjktok{中共}, \cjktok{偏股}, \cjktok{分析人士}, and \cjktok{信息仅供参考}, together with other reusable multi-character units. MinGram-PP also retains highly specific Web-derived strings and formatting artifacts, such as  
\texttt{taboola-\allowbreak interstitial-\allowbreak gallery-\allowbreak thumbnails}.
These examples illustrate the consequence of optimizing token count over a substantially larger candidate pool: long strings that occur sufficiently often in the training corpus can justify individual vocabulary entries even when they are highly specific to particular entities, templates, or source domains. Since DH-BPE uses only \(f=1.07\), many such very late merges are not present in its candidate pool in the first place; accordingly, the contrast here reflects both candidate-pool breadth and the subsequent pruning strategy.

Taken together, the three comparisons place DH-BPE between two different extremes of vocabulary construction. Relative to Pruned BPE, DH-BPE uses its modest overshoot to discover some more complete units beyond the original target boundary. Relative to MinGram, and especially to the much more heavily overshot MinGram-PP vocabulary, it more often retains reusable subword components instead of allocating vocabulary slots to fully lexicalized or highly corpus-specific sequences. The resulting vocabulary therefore does not uniformly favor either shorter or longer tokens. Rather, DP-guided exposure combined with a limited candidate overshoot produces a more selective trade-off between extending useful BPE merges and preserving compositional units. This qualitative behavior provides a complementary explanation of how DH-BPE can improve compression without requiring the substantially expanded candidate pools that allow MinGram-PP to retain large numbers of highly specialized tokens.

\section{Discussion}
\subsection{Vocabulary Construction and Parameter Selection}
\label{subsec:discussion-vocab-size}
Across the evaluated vocabulary sizes, DH-BPE consistently improves upon Standard BPE, Pruned BPE, and MinGram in the primary aggregate comparisons under the shared exact minimum-token DP encoder. More importantly, the experiments suggest that the benefit does not come from simply extending BPE training as far as possible. Relatively small overshoot
factors generally perform best, while larger candidate pools can reduce compression quality. The role of overshoot is therefore to expose enough additional merges for DP-guided selection, rather than to maximize candidate vocabulary size.

The appropriate overshoot nevertheless depends on the target vocabulary size
and pruning ratio. If the candidate pool is too small, the DP-exposure threshold
may leave too few eligible tokens to fill the final vocabulary. In practice,
this suggests using the smallest overshoot that supplies a sufficient eligible
candidate pool. The pruning-ratio experiments similarly indicate a trade-off:
more aggressive pruning can improve same-corpus compression, whereas moderate
pruning tends to provide more balanced behavior across corpora.

\subsection{Generalization Across Corpora}
\label{subsec:discussion-generalization}
A central observation is that same-corpus compression does not completely predict cross-corpus performance. Configurations that are strongest on the training distribution are not always strongest after a distribution shift, and the relative ordering of DH-BPE, MinGram, and MinGram-PP varies across Corpus III domains. This suggests that vocabulary construction involves not only maximizing compression on observed training sequences, but also deciding how much vocabulary capacity to devote to corpus-specific patterns.

The qualitative analysis provides additional insight into this trade-off. Relative to Pruned BPE, DH-BPE can use its modest overshoot to recover useful later merges that are unavailable at the original BPE target boundary. Relative to MinGram and especially heavily overshot MinGram-PP, however, DH-BPE more often retains reusable subword components rather than highly
lexicalized or corpus-specific sequences. Neither behavior is uniformly preferable: specialized long tokens can provide strong compression when their patterns recur, whereas reusable components may remain useful across a broader range of distributions. The cross-corpus results suggest that DH-BPE occupies a comparatively conservative point in this trade-off.

\subsection{Limitations}
\label{subsec:limitations}
The present study evaluates vocabulary quality primarily through encoded token
count. Although shorter encodings reduce the number of token positions required
to represent a fixed amount of text, compression alone does not establish
improvements in language-model training or downstream performance. Future work
should therefore evaluate DH-BPE within controlled language-model training to
measure effects on training loss, convergence, throughput, and downstream tasks.

The experiments also cover a limited range of corpus scales, target vocabulary
sizes, and language distributions. In addition, DH-BPE requires an extra
minimum-token DP pass over the training corpus and retains method-specific
hyperparameters for overshoot and pruning. The present work does not quantify
the resulting training-time and memory overhead or provide an automatic
parameter-selection procedure. Finally, two-space-token protection is a
targeted safeguard rather than a general treatment of whitespace and structural
formatting. Evaluating larger and more diverse corpora, measuring computational
cost, and developing more general selection criteria remain important directions
for future work.

\section{Conclusion}
This paper introduced Dynamic-Programming-Guided Hierarchical BPE (DH-BPE), a
vocabulary-construction method that combines token exposure under exact minimum-token segmentation with the dependency structure of a BPE-derived candidate vocabulary. Across the evaluated 12K and 16K settings, DH-BPE improves aggregate compression over Standard BPE, Pruned BPE, and MinGram while using only a modest BPE overshoot. MinGram-PP can achieve stronger aggregate compression when substantially larger candidate pools are used, but the cross-corpus and subset-level results show that this advantage remains domain-dependent. Together with the qualitative vocabulary analysis, the results indicate that effective vocabulary construction involves a trade-off between retaining highly specialized long tokens and preserving reusable components that remain useful across distributions. DH-BPE provides one practical mechanism for navigating this trade-off while preserving a fixed
model-visible vocabulary budget.

\end{document}